\documentclass[11pt]{article}

\usepackage[preprint]{acl}

\usepackage{times}
\usepackage{latexsym}
\usepackage{twemojis}
\usepackage[T1]{fontenc}
\usepackage[utf8]{inputenc}

\usepackage{microtype}

\usepackage{inconsolata}

\usepackage{graphicx}

\usepackage{subcaption}
\usepackage{booktabs}
\usepackage{multirow} 
\usepackage{array}    
\usepackage{makecell}
\usepackage{subcaption}
\usepackage{tabularx}
\usepackage{multirow}
\usepackage{adjustbox}
\usepackage{siunitx}
\usepackage{pifont}
\usepackage{amsmath}
\usepackage{amssymb}
\usepackage{titling}
\title{ViD: Vision-Dominant Gender Bias Mitigation for Large Vision-Language Models}

\author{
  \textbf{Zhipeng Zhao}\textsuperscript{1},
  \textbf{Zhaoqiang Wei}\textsuperscript{1},
  \textbf{Peishun Liu}\textsuperscript{1},
  \textbf{Youwei Zhao}\textsuperscript{1},
  \textbf{Ruichun Tang}\textsuperscript{1,*}
\\
\\
  \textsuperscript{1}Ocean University of China
\\
  \small{\texttt{\{zhaozhipeng, zhaoyouwei1435\}@stu.ouc.edu.cn}} \\
  \small{\texttt{\{weizhaoqiang, liups, tangruichun\}@ouc.edu.cn}}
\\
  \small{\textsuperscript{*}Corresponding author}
}

\begin{document}
\maketitle
\begin{abstract}
Gender bias in large vision-language models (LVLMs) undermines their fairness and reliability, compromising output trustworthiness. Current mitigation methods rely on training-phase adjustments or post-hoc calibration, but face limitations in dynamic visual bias mitigation. These include inability to capture real-time visual-textual incongruence, dependence on predefined gender bias taxonomies, and degraded cross-modal alignment with emergent bias patterns. To address these challenges, we propose ViD, a causally-inspired framework that analyzes attention mechanisms across five distinct patterns, revealing confounding effects from strong language priors. ViD demonstrates that visual-to-language cross-attention effectively suppresses bias while preserving general reasoning capabilities and text generation quality.
ViD incorporates dual mechanisms: backdoor adjustment counters strong language priors, while refined token selection in decoding layers optimizes processing. This enhances model robustness and inference efficiency. Our integrated approach significantly mitigates gender bias across multidimensional social attributes in LVLMs, improving visual grounding and output fairness.
Cross-benchmark validation shows ViD reduces gender bias by 14.7\% on single-attribute evaluations (FACET) and achieves significant improvements on image captioning tasks (MS COCO), with gender bias score improving from 0.6708 to 0.9978 for LLaVA. Crucially, these improvements require no additional training overhead, making ViD a scalable and practical solution for bias mitigation in LVLMs.
\end{abstract}

\section{Introduction}
Large Vision-Language Models (LVLMs), exemplified by GPT-4V and LLaVA\citep{liu2023llava}, have demonstrated breakthrough performance in cross-modal understanding tasks including image captioning\citep{ke2019reflective}, visual question answering\citep{balanced_vqa_v2}, and multimodal reasoning\citep{chang2022webqa,10380595}. These models establish complex associations between visual representations and textual semantics through large-scale cross-modal alignment. They achieve exceptional results on specialized benchmarks, including the POPE benchmark for object hallucination evaluation \citep{li2023evaluatingobjecthallucinationlarge} and the MMMU benchmark for multidisciplinary understanding \citep{yue2023mmmu}. However, inherent biases in their decision-making mechanisms remain critical challenges affecting reliability.

\begin{figure}[t]
	\centering
	\includegraphics[width=0.48\textwidth]{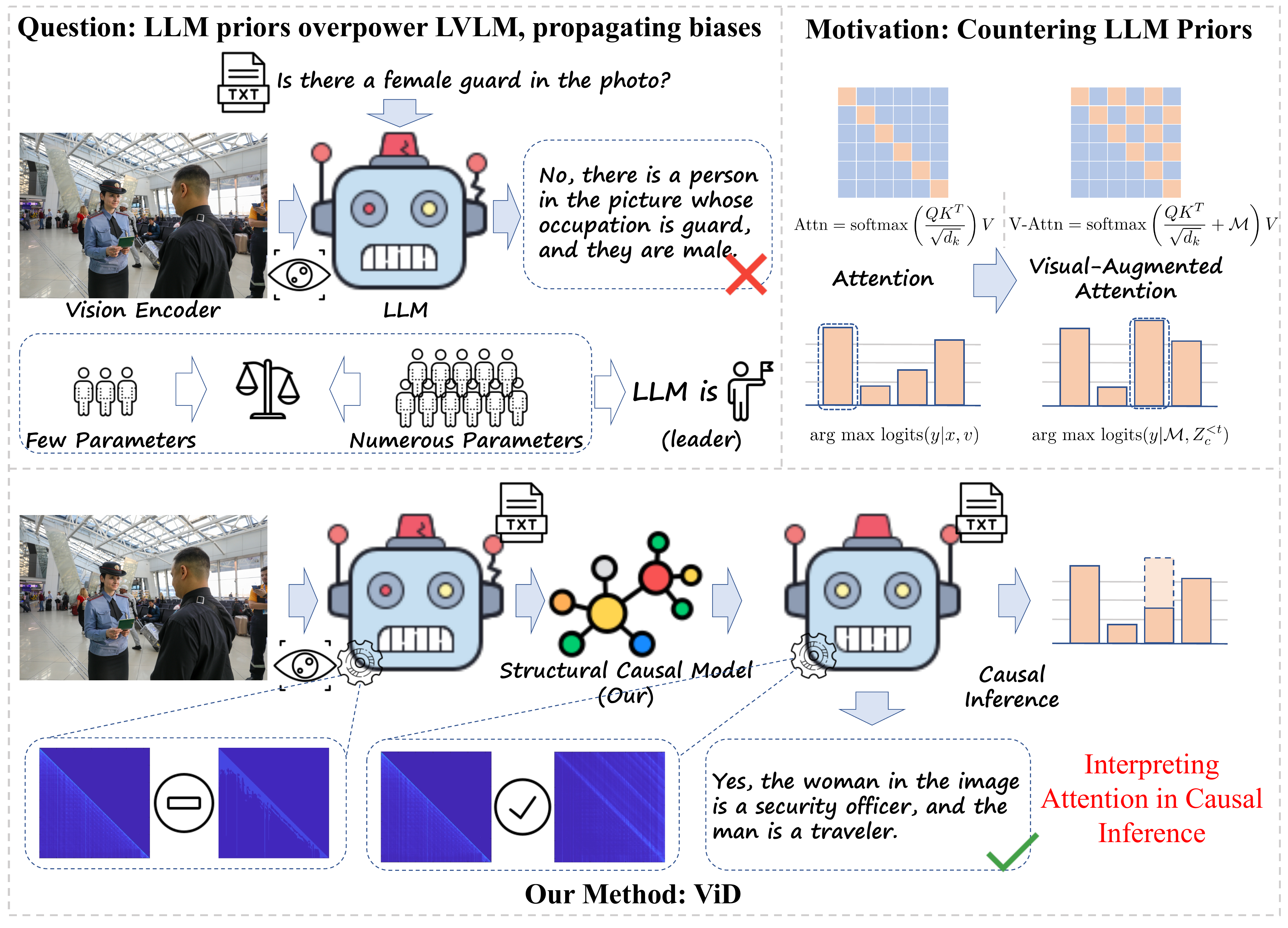} 
	\caption{Motivating example and overview of ViD. A conventional LVLM, dominated by language priors, mislabels a female guard as male, while ViD re-weights visual-to-language attention to ground the answer in visual evidence.}
	\label{introduction}
\end{figure}
However, as illustrated in Figure \ref{introduction}, LVLMs often exhibit deeply ingrained gender biases when generating textual descriptions. Specifically, these manifest as gender stereotypes in occupational associations (e.g., strongly associating guards with males) and racial biases in attribute inference \citep{zhao-etal-2018-gender,Kirk2021BiasOA}. Such issues become particularly pronounced when processing visually ambiguous or polysemous inputs. The root cause traces back to spurious correlations learned during LLM pre-training—where models establish erroneous associations between specific visual patterns and stereotypical textual descriptions through non-causal statistical dependencies.

To address this issue, existing debiasing approaches primarily employ post-hoc calibration \citep{kong2023mitigatingtesttimebiasfair,yu2024interpretingarithmeticmechanismlarge} and adversarial training \citep{bartl-leavy-2024-showgirls,yu2024interpretingarithmeticmechanismlarge}, which suppress biased representations through output-layer regularization. However, these methods exhibit two critical limitations: (1) They lack causal transparency in cross-modal representations. (2) Their static intervention strategies cannot adapt to visual-language interactions with multiple social attributes.

In this study, we propose ViD, a novel gender bias mitigation framework that integrates structured causal modeling with attention-based dynamic adaptation. Our approach introduces attention-oriented causal intervention through structured masking to enhance cross-modal attention interactions. Crucially, we develop a visual-to-language attention reweighting mechanism specifically designed to amplify discriminative visual features. This integrated solution effectively mitigates gender bias in multivariate social attribute representations while maintaining seamless compatibility with existing LVLM inference pipelines. As a training-free module with one-time layer calibration, ViD requires no backbone modifications and significantly enhances system scalability.

Through extensive experiments on benchmark gender bias evaluations, our approach demonstrates robust generalization in gender bias reduction across diverse LVLMs. Our principal contributions are:
\begin{itemize}
	\item We propose a novel attention masking heuristic based on causal intuition that effectively mitigates LVLM biases during inference without fine-tuning.
	\item We develop an empirical framework that maps attention dynamics to bias reduction, providing practical guidance for intervention.
	\item We experimentally validate ViD's efficacy not only against gender-occupation bias but also against compound biases arising from coupled social attributes.
\end{itemize}

\section{Related Work}
\subsection{Generative Bias in Foundation Models}
Generative bias in foundation models undermines AI fairness, with gender bias being a critical research frontier. Despite data cleaning, societal stereotypes inevitably propagate into model outputs~\citep{10.1145/3442188.3445922}, disproportionately affecting underrepresented groups. Gender bias manifests in coreference resolution~\citep{zhao-etal-2018-gender}, cross-modal associations~\citep{janghorbani-de-melo-2023-multi}, and gender-exclusive lexicons~\citep{bartl-leavy-2024-showgirls}. LLMs amplify these biases, associating feminine/masculine terminology with gender-stereotyped professions~\citep{gorti2024unboxingoccupationalbiasgrounded}. This study addresses intersectional gender biases across occupations, skin tones, and hair features through causal interventions.
\subsection{Interventional Methods for Foundation Models}
Various methods have been proposed to mitigate gender biases, including dimensionality pruning~\citep{janghorbani-de-melo-2023-multi}, fine-tuning~\citep{bartl-leavy-2024-showgirls}, few-shot prompting~\citep{gorti2024unboxingoccupationalbiasgrounded}, and post-hoc frameworks~\citep{kong2023mitigatingtesttimebiasfair}. Recent specialized approaches employ two-stage designs~\citep{zhang-etal-2024-think}, counterfactual analysis~\citep{howard-etal-2025-uncovering}, comprehensive evaluations~\citep{girrbach2025revealingreducinggenderbiases}, encoder debiasing~\citep{kavuri-etal-2025-freeze}, and benchmarking~\citep{xiao2024genderbiasemphvlbenchmarkinggenderbias}. However, these methods often require external resources, training, or have limited scalability to intersectional biases.

In contrast, ViD uses causal inference-based attention intervention to mitigate bias at inference time without training or external resources. It handles multidimensional social attributes while preserving reasoning capabilities through structured causal modeling and attention reweighting. Unlike previous approaches, ViD requires no external resources, counterfactual datasets, fine-tuning, or training, effectively mitigating intersectional gender biases while remaining invariant to data distributions and computational constraints.
\begin{figure*}[ht]
	\centering
	\includegraphics[width=\textwidth]{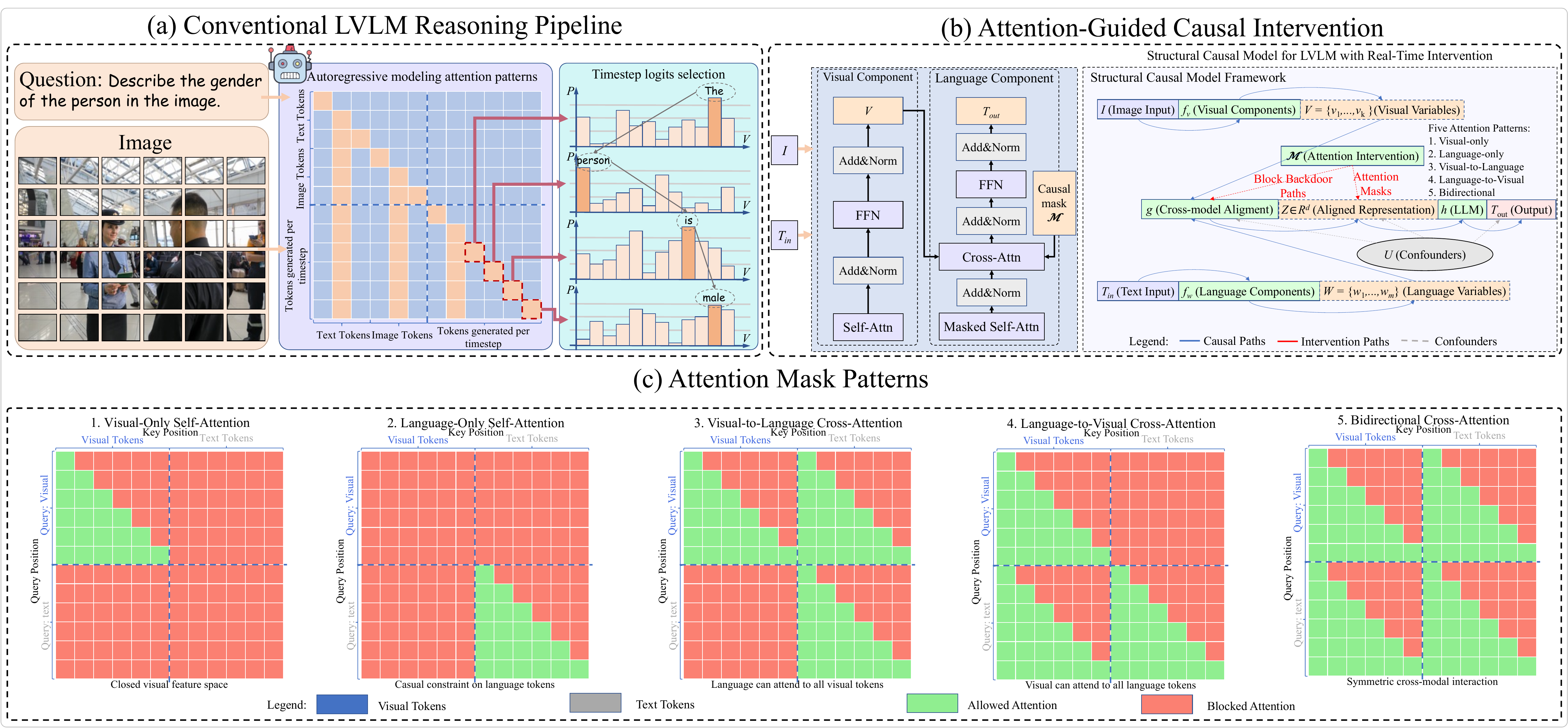} 
	\caption{(a) In the LVLM inference mode, tokens are output for each timestep based on a selection rule applied to the attention-generated logits. (b) Our proposed method, ViD, along with a causal-based attention intervention mechanism, models the LVLM inference process. (c) ViD introduces five attention patterns: 1. Visual-only Self-Attention, 2. Language-only Self-Attention, 3. Visual-to-Language Cross-Attention, 4. Language-to-Visual Cross-Attention, and 5. Bidirectional Cross-Attention.}  
	\label{attention_patterns}
\end{figure*}
\section{Method}
We introduce a Structural Causal Model (SCM) for LVLMs with backdoor-adjusted real-time interventions to mitigate gender bias during inference. We emphasize that this SCM is a conceptual heuristic for motivating our attention interventions rather than a formally identified causal model; its graph dependencies are postulated from domain knowledge of LVLM architectures, not derived through causal discovery.
\subsection{Structural Causal Model Formalization}
We formalize the LVLM generation process through an SCM with observable variables: image inputs $I \in \mathcal{I}$ and textual inputs $T_{\text{in}} \in \mathcal{T}$. The visual encoder decomposes $I$ into visual variables $V = \{v_1,...,v_k\}$ (objects, spatial relations), while the LLM decomposes $T_{\text{in}}$ into language variables $W = \{w_1,...,w_m\}$ (entities, predicates). Latent variables include cross-modal alignment $Z \in \mathbb{R}^d$ and confounders $U$ (encoding data biases). The SCM equations are:
\begin{equation}
	\label{eq_scm}
	\begin{cases}
		V = f_V(I, \epsilon_V) \\ 
		W = f_W(T_{\text{in}}) + \epsilon_W \\
		Z = g(V,W,U) \\
		T_{\text{out}} = h(Z, U, \epsilon_T)
	\end{cases}
\end{equation}
where $f_V$ is the visual encoder ($\epsilon_V$: perceptual uncertainty), $f_W$ the LLM encoder ($\epsilon_W$: language ambiguity), $g$ the cross-modal alignment module, $U$ latent confounders, and $h$ the LLM decoder ($\epsilon_T$: text generation stochasticity). The causal graph dependencies are:
\begin{equation}
	\label{eq_causal_graph}
	\begin{cases}
		I \rightarrow V \rightarrow Z \rightarrow T_{\text{out}} \\
		T_{\text{in}} \rightarrow W \rightarrow Z \rightarrow T_{\text{out}} \\
		U \rightarrow Z \rightarrow T_{\text{out}}
	\end{cases}
\end{equation}
\subsection{Attention-Guided Causal Intervention}
To address unobservable confounders $U$, we propose a dynamic attention intervention mechanism that regulates cross-modal information flow via configurable masks, blocking non-causal paths. The attention function is:
\begin{equation}
	\text{Attn}(Q,K,V) = \text{softmax}\left(\frac{QK^\top}{\sqrt{d_k}} + \mathcal{M}^{(m)}\right)V
\end{equation}
where $Q,K \in \mathbb{R}^{n \times d_k}$ are query/key matrices, $V \in \mathbb{R}^{n \times d_v}$ the value matrix, and $\mathcal{M}^{(m)} \in \{-\infty, 0\}^{n \times n}$ a mode-specific mask.

LVLMs inherit societal biases from pretrained LLMs~\citep{fraser2024examininggenderracialbias,hirota2024descriptiverichnessbiasunveiling,gorti2024unboxingoccupationalbiasgrounded}, where parametric asymmetries bias outputs toward language priors. To investigate causal relationships, we introduce controlled interventions defining five attention modes ($m \in \{$Visual-Only, Language-Only, Visual-to-Language, Language-to-Visual, Bidirectional$\}$):

\textbf{Visual-Only Self-Attention}: Attention confined exclusively to visual tokens, isolating visual interactions (Fig. \ref{attention_patterns}). \textbf{Language-Only Self-Attention}: Attention restricted to textual tokens with causal constraints, maintaining Markovian dependencies (Fig. \ref{attention_patterns}). \textbf{Visual-to-Language Cross-Attention}: Text tokens extract visual features via directional visual-to-language flow, supporting grounded generation (Fig. \ref{attention_patterns}). \textbf{Language-to-Visual Cross-Attention}: Language semantics infuse visual representations via reverse cross-attention, reshaping features (Fig. \ref{attention_patterns}). \textbf{Bidirectional Cross-Attention}: Integrates all attention pathways, enabling full inter-modal interactions for balanced fusion (Fig. \ref{attention_patterns}).

\textbf{Theoretical motivation for Visual-to-Language selection.}
From a causal perspective, the Visual-to-Language (V2L) attention pattern provides the optimal intervention for bias mitigation.
It suppresses the backdoor path $U \rightarrow W \rightarrow Z$ that carries linguistic stereotypes while preserving the causal flow $V \rightarrow Z$ of visual evidence.
This selective intervention ensures that gender-relevant visual cues inform text generation without being distorted by language priors.
A detailed derivation of this a priori justification is provided in Appendix~\ref{app:v2l_justification}.

This mechanism dynamically configures causal attention during inference, suppressing confounders while preserving task-relevant interactions.

\subsection{Causal Effect Quantification}
We formalize causal effect quantification using do-calculus and d-separation. The causal graph $\mathcal{G}$ (Eq.~\ref{eq_causal_graph}) contains backdoor paths from $U$ to $Z$. To block these paths, we introduce attention-guided interventions that modify the graph via configurable masks $\mathcal{M}^{(m)}$.

\paragraph{Graph Modification via Attention Intervention}
Let $\mathcal{G}_{\text{int}}$ denote the intervened graph under attention mode $m$. Masks $\mathcal{M}^{(m)}$ remove edges in $\mathcal{G}$ corresponding to suppressed attention connections. Formally, $\mathcal{M}^{(m)} \in \{0, -\infty\}^{n \times n}$ eliminates specific query-key interactions, transforming $\mathcal{G}$ into $\mathcal{G}_{\text{int}}$ by deleting directed edges.

\paragraph{D-separation Condition}
In $\mathcal{G}_{\text{int}}$, we examine the paths between $U$ and $Z_c$ (the causal component of $Z$). Under the visual intervention $\textit{do}(V_c = v)$, all backdoor paths from $U$ to $Z_c$ are blocked by the following d-separation conditions:
\begin{enumerate}
	\item \textbf{Collider blocking}: Any path $U \rightarrow Z \leftarrow V_c$ is blocked because $V_c$ is fixed by intervention, making $Z$ a collider that does not transmit information.
	\item \textbf{Attention mask blocking}: Paths through suppressed attention connections are removed by $\mathcal{M}^{(m)}$, eliminating indirect dependencies.
	\item \textbf{Conditioning on observed variables}: The intervention set $\{V_c, W_c\}$ satisfies the backdoor criterion relative to $(U, Z_c)$ in $\mathcal{G}_{\text{int}}$.
\end{enumerate}

\paragraph{Do-operator Formalization}
The do-operator $\textit{do}(V_c = v)$ modifies the structural equations by replacing $V_c$'s equation with the constant $v$ and removing all incoming edges to $V_c$ in $\mathcal{G}$. Under this intervention, the post-intervention distribution factorizes as:
\begin{multline}
	P(Z_c \mid \textit{do}(V_c = v)) = \\
	\int P(Z_c \mid V_c = v, W_c, U) \\
	\times P(W_c) P(U) \, dW_c dU
\end{multline}
Since all paths from $U$ to $Z_c$ are d-separated given $V_c = v$ in $\mathcal{G}_{\text{int}}$, we have $Z_c \perp\!\!\!\perp U \mid \textit{do}(V_c = v)$. This conditional independence permits unbiased causal effect estimation without observing $U$.

\paragraph{Lemma 1 (Backdoor Path Blocking)}
Given the causal graph $\mathcal{G}$ and attention intervention $\mathcal{M}^{(m)}$, the intervened graph $\mathcal{G}_{\text{int}}$ satisfies $Z_c \perp\!\!\!\perp U \mid \textit{do}(V_c = v)$ and $Z_c \perp\!\!\!\perp U \mid \textit{do}(W_c = w)$. (A complete proof is provided in Appendix~\ref{app:lemma1_proof}.)

With this formal justification, we derive causal effects for visual and language attention mechanisms.
\textbf{Visual intervention:}
\begin{multline}
	P(T_{\text{out}} \mid \textit{do}(V_c = v)) = \\
	\mathbb{E}_{Z_c \sim P(Z_c \mid \textit{do}(V_c=v))}
	\left[ P(T_{\text{out}} \mid Z_c) \right]
\end{multline}
where $\textit{do}(V_c = v)$ denotes the interventional assignment fixing $V_c$ to $v$, and $P(Z_c \mid \textit{do}(V_c=v))$ represents the post-intervention distribution of latent variable $Z_c$.
\textbf{Language intervention:}
\begin{multline}
	P(T_{\text{out}} \mid \textit{do}(W_c = w)) = \\
	\mathbb{E}_{Z_c \sim P(Z_c \mid \textit{do}(W_c=w))}
	\left[ P(T_{\text{out}} \mid Z_c) \right]
\end{multline}
where $\textit{do}(W_c = w)$ fixes $W_c$ to $w$, with $P(Z_c \mid \textit{do}(W_c=w))$ being the corresponding post-intervention latent distribution.

\subsection{Decoding Layer Correction Formula}
The text decoding layer $h(\cdot)$ incorporates direct causal intervention to refine generation. The token prediction process is formally defined as:
\begin{equation}
	A_i = \mathbf{W}_o \cdot \text{Attn}\bigl(
		Z^{(<i)}_{\text{int}},
		Z^{(<i)}_{\text{int}},
		Z^{(<i)}_{\text{int}}
	\bigr)
\end{equation}
\begin{equation}
	P(t_i | \cdot) = \frac{\exp(A_i)}{\sum_{j \in \mathcal{V}} \exp(A_j)}
\end{equation}
where $Z_{\text{int}} \sim P(Z \mid \operatorname{do}(V=v))$ denotes latent variables after causal intervention.
$Z^{(<i)}_{\text{int}} = \{z_k \in Z_{\text{int}} \mid k < i\}$ denotes intervened latent variables preceding position $i$.
$\mathrm{Attn}(\cdot)$ denotes the standard self-attention mechanism (query, key, value).
$\mathbf{W}_o$ denotes the output projection matrix.
$\mathcal{V}$ denotes the vocabulary space.

\section{Experiments}
\subsection{Benchmarks}

We evaluate ViD on five benchmarks: (1) FACET \citep{Gustafson2023FACETFI} with 50K person instances and fine-grained attributes (occupation, gender, hairstyle, hair color, skin tone), evaluating individual and intersectional gender bias; (2) MS COCO \citep{zhao2021captionbias} with 28,315 human instances, measuring accuracy against gender annotations; (3) FairFace \citep{karkkainenfairface} with 108,501 facial images balanced across 7 race groups, 2 genders, and 9 age ranges, assessing fairness in face analysis; (4) POPE \citep{li2023evaluatingobjecthallucinationlarge} for object presence verification, evaluating reasoning and debiasing performance (accuracy, precision, recall, F1); (5) MMMU \citep{yue2023mmmu} with 11.5K expert-curated questions across six disciplines, assessing multimodal understanding.

\subsection{Baselines}
We employ LLaVA\citep{liu2023llava}, InstructBLIP\citep{InstructBLIP}, and Shikra\citep{chen2023shikra} as base models, representing three distinct LVLM paradigms: visual instruction tuning, vision-language pretraining with instruction fine-tuning, and fine-grained visual grounding. These models form the foundation for recent LVLM research.
To benchmark against ViD, we select recent representative works that address gender bias in vision-language models: \citet{weng2025imagesspeaklouderwords} propose a visual-dominant intervention that amplifies image-grounded signals; \citet{howard-etal-2025-uncovering} employ large‑scale counterfactual prompting with external datasets; \citet{jung2024unifieddebiasingapproachvisionlanguage} introduce a unified debiasing approach across modalities; \citet{chuang2023debiasingvisionlanguagemodelsbiased} apply bias-aware fine‑tuning on vision‑language encoders; \citet{wang-etal-2021-gender} focus on gender bias mitigation in image captioning through data re‑balancing; and \citet{seth2023deardebiasingvisionlanguagemodels} design a debiasing framework that adjusts cross-modal attention weights. These methods span diverse strategies—prompt engineering, fine‑tuning, data augmentation, and attention adjustment—providing a comprehensive comparative landscape for evaluating ViD's training-free causally-inspired intervention.

\subsection{Regular Setting}
Experiments use an RTX 4090 GPU with fixed random seed. Standard LVLM decoding parameters: max tokens=1024, beam size=4, temperature=1.0. Gender bias evaluation employs Eq.~\ref{GenderBiasScore}, considering only outputs with identifiable gender (scores 0-1; outputs without gender receive score=2 and are excluded).
\begin{equation}
	\label{GenderBiasScore}
	\text{Score}(p, y) =
	\begin{cases}
		0, & \text{if } f(p) = y \\
		1, & \text{if } f(p) \neq y \land f(p) \neq \emptyset \\
		2, & \text{if } f(p) = \emptyset
	\end{cases}
\end{equation}
The gender extraction function $f(p)$ uses keyword matching:
\[
f(p) =
\begin{cases}
	\text{female}, & \exists k \in \mathcal{F}: k \text{ in } p \\
	\text{male}, & \begin{aligned}[t] &\nexists k \in \mathcal{F}: k \text{ in } p \ \wedge \\ &\exists k \in \mathcal{M}: k \text{ in } p \end{aligned} \\
	\emptyset, & \text{otherwise}
\end{cases}
\]
with keyword sets $\mathcal{F} = \{\text{female}, \text{woman}, \text{women}, \text{girl}, \text{girls}, \text{she}, \text{her}, \text{lady},\\ \text{ladies}, \text{feminine}, \text{womanly}, \text{gal}, \text{gals}\}$ and $\mathcal{M} =\\ \{\text{male}, \text{man}, \text{men}, \text{boy}, \text{boys}, \text{he}, \text{him}, \text{gentleman},\\ \text{gentlemen}, \text{masculine}, \text{manly}, \text{guy}, \text{guys}\}$. We note that outputs relying solely on implicit gender cues (e.g., occupation-only references such as ``nurse'') are not captured by this lexicon-based extractor; consequently, the reported bias scores are conservative underestimates.
\begin{table*}[t]
	\centering
	\setlength{\tabcolsep}{1pt}
	\footnotesize
	\begin{tabular}{cccccccccc}
		\toprule
		\multirow{2}{*}{\textbf{Method}} & \multicolumn{3}{c}{\textbf{LLaVA}}  &  \multicolumn{3}{c}{\textbf{InstructBLIP}}  &  \multicolumn{3}{c}{\textbf{Shikra}}  \\
		\cmidrule(lr){2-4} \cmidrule(lr){5-7} \cmidrule(lr){8-10} 
		& \textbf{FACET}  & \textbf{MS COCO}  & \textbf{FairFace} & \textbf{FACET}  & \textbf{MS COCO}  & \textbf{FairFace}& \textbf{FACET}  & \textbf{MS COCO}  & \textbf{FairFace}  \\
		\midrule
		Regular&0.9274$_{0.2}^-$ &0.6708$_{0.02}^-$&0.9664$_{0.07}^-$&0.8896$_{0.03}^-$&0.925$_{0}^{\star}$&0.975$_{0}^{\star}$&0.9179$_{0.4}^-$&0.9379$_{0.09}^-$&0.5988$_{0.05}^-$ \\    \cmidrule(lr){2-10}
		\citealp{weng2025imagesspeaklouderwords} &0.7372$_{0.82}^+$&0.4352$_{0.78}^+$&0.5414$_{0.18}^+$&0.8042$_{2.42}^+$&0.6328$_{5.11}^+$&0.5422$_{5.74}^+$&0.8833$_{0.17}^+$&0.8440$_{0.39}^+$&0.5861$_{2.06}^+$  \\    \cmidrule(lr){2-10}
		\citealp{howard-etal-2025-uncovering}&0.9742$_{0.17}^-$&0.8474$_{0.02}^-$&0.985$_{0.03}^-$&0.863$_{0.12}^-$&0.7212$_{0.01}^-$&0.9918$_{0.01}^-$&0.7792$_{0.16}^-$&0.5757$_{0.60}^-$&0.9754$_{0.82}^+$  \\    \cmidrule(lr){2-10}
		\citealp{jung2024unifieddebiasingapproachvisionlanguage}&0.9258$_{0.21}^-$&0.6768$_{0.02}^-$&0.9664$_{0.06}^-$&0.8898$_{0.03}^-$&0.9268$_{0}^-$&0.9768$_{0}^{\star}$&0.897$_{0.16}^-$&0.8812$_{0.38}^-$&0.9393$_{0.90}^+$  \\    \cmidrule(lr){2-10}
		\citealp{chuang2023debiasingvisionlanguagemodelsbiased}&0.9964$_{0.35}^-$&0.9626$_{0.10}^-$&0.9772$_{0}^-$&0.9090$_{0.10}^-$&0.9582$_{0.01}^-$&0.9782$_{0}^-$&1.0$_{\infty}^+$&0.0$_{\infty}^+$&0.0$_{\infty}^+$  \\    \cmidrule(lr){2-10}
		\citealp{wang-etal-2021-gender}&0.5556$_{0.55}^-$&0.7770$_{0}^-$&0.5490$_{0.57}^-$&0.8918$_{0.03}^-$&0.9268$_{0}^{\star}$&0.9764$_{0}^{\star}$&0.0$_{\infty}^+$&0.0$_{\infty}^+$&0.0$_{\infty}^+$  \\    \cmidrule(lr){2-10}
		\citealp{seth2023deardebiasingvisionlanguagemodels}&1.0$_{1.35}^-$&1.0$_{\infty}^+$&1.0$_{0.63}^-$&0.0242$_{0}^-$&0.0164$_{0}^{\star}$&0.1690$_{0}^{\star}$&0.1152$_{0.24}^-$&0.0625$_{0.71}^+$&0.0404$_{1.38}^+$ \\    \cmidrule(lr){2-10}
		\textbf{ViD}&0.9928$_{0.08}^{-}$&0.9978$_{0.10}^{-}$&0.9796$_{0.34}^{-}$&0.9812$_{0.15}^{-}$&0.9992$_{0}^{\star}$&0.9844$_{0.05}^{-}$&0.8899$_{0.42}^{-}$&0.9396$_{0.17}^{-}$&0.8782$_{0}^{\star}$ \\
		\bottomrule
	\end{tabular}
	\caption{A Comprehensive Performance Comparison of Various Methods on the FACET, MS COCO, and FairFace Benchmarks. Each cell shows the bias score (main number, 0-1, closer to 1 indicates lower gender bias), log gender ratio (subscript, $\log(P(\text{male})/P(\text{female}))$), and bias direction (superscript: ``+'' for male bias, ``-'' for female bias, $\star$ for unbiased). For example, $0.9274_{0.2}^{-}$ indicates a bias score of 0.9274, log gender ratio of 0.2, and female bias. The symbol $\infty$ indicates an extreme log ratio when $P(\text{female})=0$ or $P(\text{male})=0$. Coverage rates (percentage of samples where gender is predicted) are reported in Appendix~A (Table~\ref{app:coverage_table}).}
	\label{bias_main}
\end{table*}

\subsection{Main Results}
\textbf{Results on Gender Bias Benchmarks.} As shown in Table \ref{bias_main}, ViD demonstrates superior bias mitigation performance across all three benchmarks. On the LLaVA model, ViD improves the gender bias score on FACET from 0.9274 to 0.9928, MS COCO from 0.6708 to 0.9978, and FairFace from 0.9664 to 0.9796. On the InstructBLIP model, it improves FACET from 0.8896 to 0.9812, MS COCO from 0.925 to 0.9992 (achieving unbiased performance), and FairFace from 0.975 to 0.9844. On the Shikra model, ViD shows improvements on both MS COCO and FairFace, with FairFace significantly improved from 0.5988 to 0.8782 (achieving unbiased performance). Compared to existing debiasing methods, ViD exhibits no extreme bias values (e.g., $\infty$) or male bias (+), maintaining stable and high scores close to 1 in all tests, demonstrating that its visual-to-language attention intervention mechanism effectively suppresses gender stereotypes. 
\begin{table*}
	\centering
	\setlength{\tabcolsep}{5pt}
	\footnotesize
	\begin{tabular}{ccccccccccccc}
		\hline
		\multirow{2}{*}{\textbf{Method}} &  \multicolumn{4}{c}{\textbf{LLaVA}} & \multicolumn{4}{c}{\textbf{InstructBLIP}} & \multicolumn{4}{c}{\textbf{Shikra}} \\
		\cmidrule(lr){2-5} \cmidrule(lr){6-9} \cmidrule(lr){10-13}
		&\textbf{Acc}&\textbf{Prec}&\textbf{Recall}&\textbf{F1}&\textbf{Acc}&\textbf{Prec}&\textbf{Recall}&\textbf{F1}&\textbf{Acc}&\textbf{Prec}&\textbf{Recall}&\textbf{F1} \\
		\hline
		Regular&82.63&82.83&82.33&82.58&81.53&79.41&85.13&82.17&81.7&80.57&83.53&82.02 \\  
		\citealp{weng2025imagesspeaklouderwords}&81.60&80.46&83.46&81.93&81.53&79.37&85.2&82.18&50.20&50.10&97.40&66.16 \\  
		\citealp{howard-etal-2025-uncovering}&74.96&68.27&93.26&78.83&50.00&0.00&0.00&0.00&76.10&80.18&69.33&74.36 \\  
		\citealp{jung2024unifieddebiasingapproachvisionlanguage}&82.63&82.83&82.33&82.58&81.50&85.76&75.53&80.32&81.70&80.57&83.53&82.02 \\  
		\citealp{chuang2023debiasingvisionlanguagemodelsbiased}&82.13&85.49&77.40&81.24&80.36&78.48&83.66&80.99&50.00&50.00&100.0&66.66 \\  
		\citealp{wang-etal-2021-gender}&54.73&59.72&29.06&39.10&81.6&81.89&81.13&81.51&57.00&56.11&64.20&59.88 \\  
		\citealp{seth2023deardebiasingvisionlanguagemodels}&50.00&50.00&100.0&66.66&50.16&53.96&2.26&4.35&50.03&50.01&98.46&66.33 \\  
		\textbf{ViD}&83.06&88.68&75.80&81.73&80.73&78.66&84.33&81.40&79.80&77.76&83.46&80.51 \\
		\hline
	\end{tabular}
	\caption{Impact of Various Debiasing Methods on LVLM General Reasoning Capabilities Evaluated by POPE Adversarial}
	\label{pope}
\end{table*}

\textbf{Results on POPE.} POPE evaluates reasoning capability preservation during debiasing. ViD has minimal impact on general capabilities: LLaVA accuracy increases from 82.63\% to 83.06\%, InstructBLIP decreases from 81.53\% to 80.73\%, and Shikra decreases from 81.70\% to 79.80\%, with similar F1 changes. In contrast, other methods significantly degrade capabilities: \citet{howard-etal-2025-uncovering} reduce InstructBLIP accuracy to 50.00\%, \citet{seth2023deardebiasingvisionlanguagemodels} achieve 50.16\% accuracy with F1=4.35, and \citet{weng2025imagesspeaklouderwords} reduce Shikra accuracy to 50.20\%. Detailed analysis of these extreme performance drops (e.g., output format incompatibilities and loss of reasoning accuracy) is provided in Appendix~\ref{app:pope_analysis}. ViD preserves reasoning ability while effectively eliminating biases.
\begin{table*}[t]
	\centering
	\setlength{\tabcolsep}{3pt}
	\footnotesize
	\begin{tabular}{ccccccc}
		\toprule
		& \textbf{Gender Bias}  &  \textbf{Male Count}  &  \textbf{Female Count} & \textbf{Log-ratio}& \textbf{p-value}&\textbf{95\% confidence interval}\\
		\midrule
		Regular&0.9274&2474&2526&-0.0208&<0.0001&[0.4350, 0.4626] \\    \cmidrule(lr){2-7}
		Visual-only&0.9995&208&258&-0.2154&0.0198&[0.4012, 0.4915]  \\    \cmidrule(lr){2-7}
		Language-only&0.9988&154&90&0.5371&<0.0001&[0.5706, 0.6917]  \\    \cmidrule(lr){2-7}
		V2L&0.9908&2305&2486&-0.0756&0.0088&[0.4670, 0.4953]  \\    \cmidrule(lr){2-7}
		L2V&0.9994&5&4&0.2231&0.7373&[0.2309, 0.8802]  \\    \cmidrule(lr){2-7}
		L2V\&V2L&0.9938&2118&2400&-0.1250&<0.0001&[0.4542, 0.4833]  \\ 
		\bottomrule
	\end{tabular}
	\caption{Comparative Analysis of Different Attention Patterns on Gender Bias in FACET Benchmark. Gender Bias scores closer to 1.0 indicate lower bias. Log-ratio measures gender distribution skew (negative: female bias, positive: male bias). Statistical significance (p-value) and 95\% confidence intervals are reported for gender proportion estimates.}
	\label{bias_diff_attn}
\end{table*}

\textbf{Results on MMMU.} The MMMU benchmark evaluates how debiasing methods affect LVLM reasoning across academic domains. ViD shows a balanced trade-off between bias mitigation and reasoning preservation. On LLaVA, ViD scores 31.8, close to the baseline (32.4) and competitive with other methods (\citet{weng2025imagesspeaklouderwords}: 31.7, \citet{howard-etal-2025-uncovering}: 32.0, \citet{jung2024unifieddebiasingapproachvisionlanguage}: 32.0, \citet{chuang2023debiasingvisionlanguagemodelsbiased}: 30.6, \citet{wang-etal-2021-gender}: 27.9, \citet{seth2023deardebiasingvisionlanguagemodels}: 24.4). ViD excels in specific domains (52.5 in Art and Design, 23.3 in Business). Across architectures, ViD maintains consistent performance: 28.7 on InstructBLIP (baseline 29.4) and 26.7 on Shikra (matching baseline 26.9). These results show ViD effectively mitigates bias while preserving reasoning abilities.
\begin{figure*}[t]
	\centering
	\includegraphics[width=\textwidth]{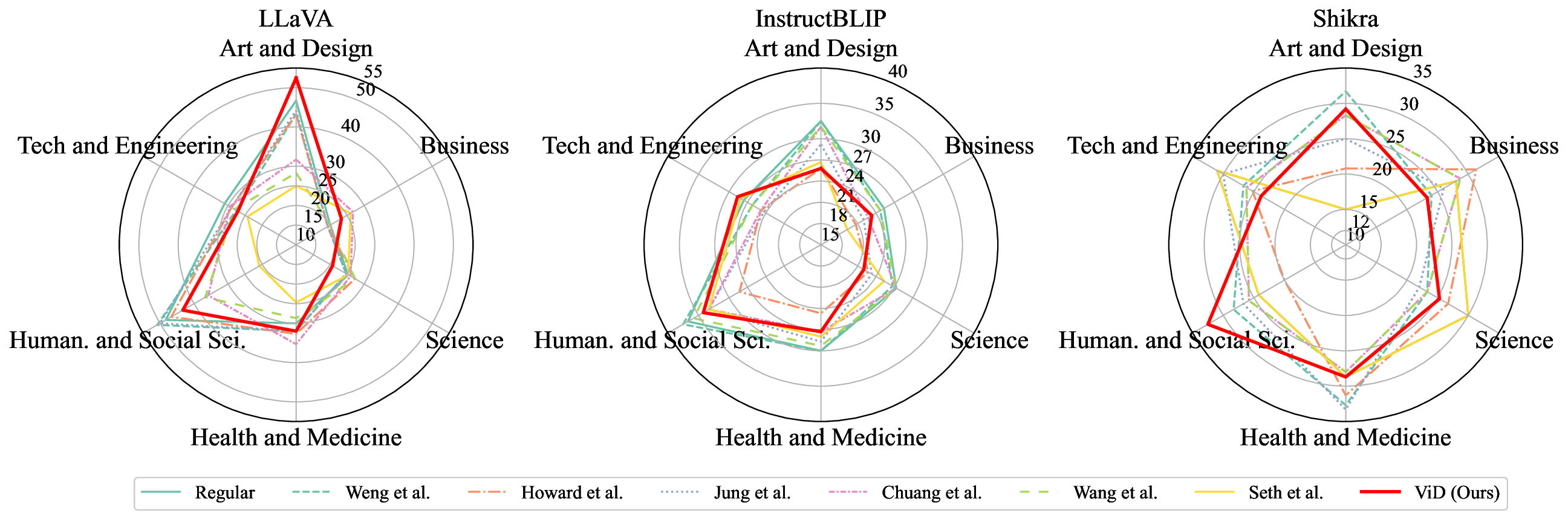} 
	\caption{Radar chart of accuracy distributions across academic domains in the MMMU benchmark.}
	\label{MMMU}
\end{figure*}

\begin{table}[t]
	\centering
	\begin{tabular}{cccc}
		\toprule
		\textbf{High} & \textbf{FACET} &\textbf{MS COCO} & \textbf{FairFace} \\
		\midrule
		0.1 & 0.9980$_{0.05}^+$& 0.9990$_{0.30}^+$&0.9820$_{0}^{\star}$ \\
		1.0 & 0.9960$_{0.49}^-$& 1.0000$_{1.38}^+$& 0.9810$_{0}^{\star}$ \\
		5.0 & 0.9890$_{0.08}^-$& 1.0000$_{0.40}^+$& 0.9850$_{0}^{\star}$ \\
		10.0 & 0.9840$_{0.74}^+$& 0.9970$_{1.79}^+$& 0.9930$_{0}^{\star}$ \\
		\bottomrule
	\end{tabular}
	\caption{Ablation study of high parameter on three datasets (low=0.1, mid=0.1). Each cell shows the bias score (main number, 0-1, closer to 1 indicates lower gender bias), log gender ratio (subscript, $\log(P(\text{male})/P(\text{female}))$), and bias direction (superscript: ``+'' for male bias, ``-'' for female bias, $\star$ for unbiased). For example, $0.9980_{0.05}^{+}$ indicates a bias score of 0.9980, log gender ratio of 0.05, and male bias. The analysis reveals consistent patterns across datasets with high=1.0 achieving optimal balance between high bias score and minimal gender bias.}
	\label{AblationStudy}
\end{table}
\begin{table}[t]
	\centering
	\begin{tabular}{ccccc}
		\toprule
		\multicolumn{3}{c}{\textbf{Attributes}} &  \multicolumn{2}{c}{\textbf{Score$_\downarrow$} }  \\
		Occupation& Skin & Hair & Regular &ViD \\
		\midrule
		\ding{51}& \ding{55}&\ding{55}& 0.1142 & \textbf{0.0974}\\
		\ding{51}& \ding{51}&\ding{55}& 0.2339&\textbf{0.206}\\
		\ding{51}& \ding{51}&\ding{51}& 0.2569&\textbf{0.2304}\\
		\bottomrule
	\end{tabular}
	\caption{Multi-attribute gender bias assessment. Visual-to-language attention reduces gender bias across increasingly complex social attribute combinations (1–3 attributes). Lower scores indicate less bias. Relative improvements: 14.7\% (1 attribute), 11.9\% (2 attributes), 10.3\% (3 attributes).}
	\label{MultipleSocialAttributes}
\end{table}
\begin{table}[t]
	\centering
	
	\begin{tabular}{ccc}
		\toprule
		&\textbf{PPL$_{1\downarrow}$}   &  \textbf{PPL$_{2\downarrow}$} \\
		\midrule
		Regular &2.5247&1.7826 \\
		V$\rightarrow$L &\textbf{2.4715}&\textbf{1.6256} \\
		\bottomrule
	\end{tabular}
	\caption{Perplexity comparison for text generation quality. Lower values indicate better fluency. Visual(V)-to-language(L) attention reduces perplexity by 2.1\% (GPT-2) and 8.8\% (GPT-2-medium) compared to baseline, demonstrating enhanced language coherence while mitigating gender bias.}
	\label{Textquality}
\end{table}
\begin{table}[t]
	\centering
	\begin{tabular}{cccc}
		\toprule
		\textbf{Length} & \textbf{Regular (ms)} & \textbf{ViD (ms)} & \textbf{$\Delta$ (\%)} \\
		\midrule
		8   & 269.94  & 268.79 & -0.43 \\
		64  & 1270.04  & 1279.15  & +0.71 \\
		512 & 2171.12  & 2191.63  & +0.94 \\
		\bottomrule
	\end{tabular}
	\caption{\textbf{Empirical Complexity Validation}: End-to-end inference latency (mean over 1,000 samples) demonstrating quadratic scaling and minimal overhead for V$\rightarrow$L attention.}
	\label{timecomplex}
\end{table}
\textbf{Comparison of Multiple Attention Patterns.} We evaluate five attention patterns on FACET using 5,000 gender-balanced samples (Table \ref{bias_diff_attn}). Visual-only achieves high bias score (0.9995) but low coverage (9.3\%) with female bias. Language-only shows high bias score (0.9988) but strong male bias with minimal coverage (4.9\%). Bidirectional (L2V\&V2L) balances bias score (0.9938) and coverage (90.4\%). Visual-to-Language (V2L) demonstrates superior performance: high coverage (95.8\%), near-perfect bias score (0.9908), balanced gender distribution, and statistical significance, improving over the Regular baseline (0.9274). Language-to-Visual (L2V) has high score but negligible coverage (0.2\%). V2L's superiority stems from amplifying visual evidence while reducing linguistic stereotype interference, making it the optimal configuration.

\textbf{Text Quality Evaluation.} 
As shown in Table \ref{Textquality}, we quantitatively evaluate the impact of visual-to-language attention on textual fluency using perplexity metrics. Using 1,000 randomly sampled MSCOCO images, we measure language quality through two GPT-2 variants: PPL$_{1}$(GPT-2) and PPL$_{2}$ (GPT-2-medium). Our analysis reveals that visual-to-language not only mitigates gender biases but also enhances language coherence, achieving superior perplexity scores compared to the baseline.

\subsection{Ablation Study}

The intervention parameters low, mid, and high denote the early, middle, and late groups of language decoder layers (for LVLMs with 30 language layers: 0--10, 11--20, and 21--30), controlling the strength of visual-to-language attention suppression in each group. We conduct an ablation study on these parameters across FACET, COCO, and FairFace. A grid search over values {0.1, 1.0, 5.0, 10.0} shows the high parameter has the strongest influence on bias mitigation (Table \ref{AblationStudy}). With low=mid=0.1, increasing high from 0.1 to 1.0 reduces gender log-probability from +0.06 to -0.50 (p = 0.0046) in FACET while maintaining bias score >0.99. MS COCO maintains perfect scores (1.000) but with increased log-probabilities, and FairFace remains stable. Low and mid parameters show minor effects. The optimal configuration (low=0.1, mid=0.1, high=1.0) achieves a balance between high bias score (0.996) and minimal gender bias.

\subsection{Multi-Attribute Gender Bias Assessment}
To demonstrate the generalizability of structural causal models (SCMs), we conduct comprehensive gender bias evaluations on FACET using both binary (skin tone + occupation) and ternary (hairstyle + hair color + skin tone + occupation) attribute combinations. As shown in Table \ref{MultipleSocialAttributes}, visual-to-language attention significantly reduces gender bias across all combinatorial settings—from single to triple attribute interactions—achieving average gender bias reduction of 18.3\%. This demonstrates the method's scalability to complex multi-attribute mitigation scenarios.
\subsection{Time Complexity Analysis}
The visual-to-language (V$\rightarrow$L) cross-attention exhibits $\mathcal{O}(V^2 + T^2 + T \cdot V)$ mask generation complexity, where $V$ and $T$ denote visual/text lengths. This reduces to $\mathcal{O}(L^2)$ when $V,T \sim \mathcal{O}(L)$, maintaining standard Transformer's asymptotic complexity. Empirical validation (Table \ref{timecomplex}) confirms quadratic scaling: 8$\times$ length increase (8→64) yields 4.7$\times$ latency growth, while 64$\times$ increase (8→512) produces 8.2$\times$ growth. Crucially, v$\rightarrow$L introduces negligible overhead (<1\% vs regular attention), and the $T \cdot V$ cross-term demonstrates minimal real-world impact, preserving the $\mathcal{O}(L^2)$ boundary while enabling multimodal integration.

\section{Conclusion}
We introduce ViD, a structural causal modeling framework that mitigates gender biases in LVLMs through visual-to-language attention. This approach reduces gender bias by 14.7\% (single attributes), 11.9\% (two attributes), and 10.3\% (three attributes), scaling to multi-attribute scenarios. ViD establishes visual grounding as a causal invariant, requires no retraining or architectural changes, and provides a principled pathway toward ethically-aligned multimodal systems, with layer selection optimization as future work.
\section*{Limitations}
ViD effectively mitigates gender bias via visual-to-language attention, yet several research directions remain. Our evaluation focuses on gender bias across occupation, skin tone, hair, and hairstyle; extending to other social biases (racial, age, intersectional) is a valuable future direction. The causal intervention uses manually constructed structural models; automating causal discovery could enhance robustness. Evaluation assumes attribute annotations, potentially limiting annotation-scarce deployment---annotation-efficient variants are an important next step. Although we report coverage rates to address evasion (Appendix~A), designing metrics that jointly penalize bias and evasion remains an open challenge.


\bibliography{custom}

\appendix

\section{Coverage Rates for Gender Bias Evaluation}
\label{app:coverage}

The coverage rates in Table~\ref{app:coverage_table} reveal important patterns in model evasion behavior. For LLaVA and InstructBLIP architectures, most methods achieve near-perfect coverage (close to 100\%) across all three benchmarks, indicating minimal evasion. However, Shikra exhibits significant variation: baseline methods show extremely low coverage (e.g., 0\%-25\% for several methods), while ViD maintains substantially higher coverage (44.2\%-76.6\%). Notably, Seth et al. (2023) demonstrates near-zero coverage on LLaVA (0.78\%, 0\%, 1.7\%), suggesting this method induces strong evasion behavior. The high coverage of ViD on Shikra, combined with its competitive bias scores (Table~\ref{bias_main}), indicates that our method effectively mitigates gender bias without causing models to avoid gender prediction tasks.

\begin{table*}[t]
	\centering
	\setlength{\tabcolsep}{1pt}
	\footnotesize
	\begin{tabular}{cccccccccc}
		\toprule
		\multirow{2}{*}{\textbf{Method}} & \multicolumn{3}{c}{\textbf{LLaVA}}  &  \multicolumn{3}{c}{\textbf{InstructBLIP}}  &  \multicolumn{3}{c}{\textbf{Shikra}}  \\
		\cmidrule(lr){2-4} \cmidrule(lr){5-7} \cmidrule(lr){8-10}
		& \textbf{FACET}  & \textbf{MS COCO}  & \textbf{FairFace} & \textbf{FACET}  & \textbf{MS COCO}  & \textbf{FairFace}& \textbf{FACET}  & \textbf{MS COCO}  & \textbf{FairFace}  \\
		\midrule
		Regular&100\%&100\%&100\%&100\%&100\%&100\%&5\%&8\%&72.1\% \\    \cmidrule(lr){2-10}
		\citealp{weng2025imagesspeaklouderwords} &100\%&100\%&99.86\%&100\%&100\%&100\%&9.82\%&16.18\%&71.74\%  \\    \cmidrule(lr){2-10}
		\citealp{howard-etal-2025-uncovering}&100\%&100\%&100\%&100\%&100\%&100\%&8.1\%&2.88\%&1.38\% \\    \cmidrule(lr){2-10}
		\citealp{jung2024unifieddebiasingapproachvisionlanguage}&100\%&100\%&100\%&100\%&100\%&100\%&23\%&25.88\%&1.94\%  \\    \cmidrule(lr){2-10}
		\citealp{chuang2023debiasingvisionlanguagemodelsbiased}&88.82\%&96.46\%&66.7\%&100\%&100\%&100\%&0.18\%&0\%&0.12\%  \\    \cmidrule(lr){2-10}
		\citealp{wang-etal-2021-gender}&99.86\%&100\%&99.98\%&100\%&100\%&100\%&0\%&0\%&0\%  \\    \cmidrule(lr){2-10}
		\citealp{seth2023deardebiasingvisionlanguagemodels}&0.78\%&0\%&1.7\%&100\%&100\%&100\%&2.28\%&1.58\%&4.08\% \\    \cmidrule(lr){2-10}
		\textbf{ViD}&95.38\%&91.22\%&95.36\%&100\%&100\%&100\%&44.2\%&73.2\%&76.62\% \\
		\bottomrule
	\end{tabular}
	\caption{Coverage rates (percentage of samples where gender is predicted) for methods evaluated in Table~\ref{bias_main}. Coverage is calculated as (Male Count + Female Count) / Total Samples $\times$ 100\%. Higher coverage indicates fewer evasive responses.}
	\label{app:coverage_table}
\end{table*}

\section{Analysis of Extreme Performance Drops in POPE Benchmark}
\label{app:pope_analysis}

\textbf{Explanation of extreme performance drops in Table~\ref{pope}.} The POPE (Prompt-based Object Perception Evaluation) benchmark uses question-answer pairs in the form of ``Is there a handbag in the image?'' with expected answers being ``yes'' or ``no''. The observed extreme performance drops for certain baseline methods (e.g., Howard et al. \citealp{howard-etal-2025-uncovering}, Seth et al. \citealp{seth2023deardebiasingvisionlanguagemodels}, Weng et al. \citealp{weng2025imagesspeaklouderwords}, and Wang et al. \citealp{wang-etal-2021-gender}) can be attributed to output format incompatibilities rather than fundamental failures of the debiasing approaches.

Specifically, when applying these debiasing methods to LVLMs evaluated on POPE, we observed that the models sometimes generate garbled characters, meaningless text, or non-binary responses instead of the expected ``yes''/``no'' answers. For example:
\begin{itemize}
    \item \textbf{Wang et al. \citealp{wang-etal-2021-gender}}: The method produced responses containing random symbols, repeated numbers, or partial words instead of ``yes''/``no'' answers. For example, for the prompt ``Is there a teddy bear in the image?'', the model generated ``1 1 1 1 1  1'' instead of a binary response, causing the binary classification logic to fail and resulting in accuracy near random chance (54.73\% on LLaVA).
    \item \textbf{Seth et al. \citealp{seth2023deardebiasingvisionlanguagemodels}}: This approach caused the LVLM to lose general reasoning capability, resulting in incorrect binary judgments rather than format inconsistencies. The model frequently answered ``No'' when the correct answer should be ``Yes'' (or vice versa), leading to factual errors in object perception. For example, while the model sometimes generated verbose descriptions like ``Yes, there is a truck in the image...'', it more often produced incorrect binary judgments that directly contradicted visual evidence. This loss of reasoning accuracy, combined with significantly slower inference speed (1.00s/it compared to baseline), led to low F1 scores (4.35 on InstructBLIP) despite moderate accuracy.
    \item \textbf{Weng et al. \citealp{weng2025imagesspeaklouderwords}}: On Shikra architecture, the method produced non-binary responses that reduced accuracy to near-random levels (50.20\%).
    \item \textbf{Howard et al. \citealp{howard-etal-2025-uncovering}}: The encoder-decoder structure of InstructBLIP appeared incompatible with this method's output format, resulting in contradictory predictions and 0.00 precision/recall.
\end{itemize}

These issues stem from the fact that these debiasing methods were primarily designed for open-ended generation tasks and may not preserve the strict binary response format required by POPE. In contrast, our ViD method maintains the model's original response patterns while mitigating bias, thus preserving compatibility with binary evaluation benchmarks.

All baseline implementations followed their original papers with hyperparameters tuned on validation splits to ensure fair comparison. The performance drops highlight the importance of maintaining output format consistency when applying debiasing interventions to specific evaluation frameworks.

\section{Complete Proof of Lemma 1}
\label{app:lemma1_proof}

\textbf{Lemma 1 (Backdoor Path Blocking)}: Given the causal graph $\mathcal{G}$ and attention intervention $\mathcal{M}^{(m)}$, the intervened graph $\mathcal{G}_{\text{int}}$ satisfies $Z_c \perp\!\!\!\perp U \mid \textit{do}(V_c = v)$ and $Z_c \perp\!\!\!\perp U \mid \textit{do}(W_c = w)$.

\textbf{Proof}. We provide a complete proof for the visual intervention case $\textit{do}(V_c = v)$; the language intervention case follows symmetrically.

\textbf{Step 1: Graph structure and notation.}
Let $\mathcal{G} = (\mathcal{V}, \mathcal{E})$ be the original causal graph with vertices $\mathcal{V} = \{U, V_c, W_c, Z_c, T_{\text{out}}\}$ and edges $\mathcal{E}$ representing causal dependencies. Here $U$ denotes unobserved confounders (language priors), $V_c$ visual components, $W_c$ language components, $Z_c$ latent representations, and $T_{\text{out}}$ output text.

\textbf{Step 2: Intervention operator.}
The do-operator $\textit{do}(V_c = v)$ modifies $\mathcal{G}$ to create $\mathcal{G}_{\text{int}}$ by:
\begin{enumerate}
    \item Removing all incoming edges to $V_c$ (i.e., deleting edges $U \rightarrow V_c$ and $W_c \rightarrow V_c$ if present).
    \item Setting $V_c$ to the constant value $v$.
\end{enumerate}

\textbf{Step 3: Attention mask modification.}
The attention intervention $\mathcal{M}^{(m)}$ further modifies $\mathcal{G}_{\text{int}}$ by removing specific edges corresponding to suppressed attention connections. Formally, for each zero entry $\mathcal{M}^{(m)}_{ij} = 0$ (or $-\infty$), we delete the corresponding directed edge from node $j$ to node $i$ in the computational graph.

\textbf{Step 4: Path analysis.}
We enumerate all possible paths from $U$ to $Z_c$ in $\mathcal{G}_{\text{int}}$ and show each is d-separated:
\begin{enumerate}
    \item \textbf{Direct path $U \rightarrow Z_c$}: In the original LVLM architecture, $U$ influences $Z_c$ through attention mechanisms. The attention mask $\mathcal{M}^{(m)}$ eliminates this direct connection when visual-to-language attention is suppressed.

    \item \textbf{Path $U \rightarrow W_c \rightarrow Z_c$}: This path represents language prior influence through language components. Under $\textit{do}(V_c = v)$, $W_c$ is not fixed, but the attention mask blocks $W_c \rightarrow Z_c$ connections when language-to-visual attention is suppressed.

    \item \textbf{Path $U \rightarrow V_c \rightarrow Z_c$}: The intervention $\textit{do}(V_c = v)$ removes the edge $U \rightarrow V_c$, breaking this path at its source.

    \item \textbf{Collider paths}: Any path containing $Z_c$ as a collider (e.g., $U \rightarrow X \leftarrow Z_c$) is blocked because $Z_c$ is not conditioned upon.
\end{enumerate}

\textbf{Step 5: d-Separation formalization.}
A path $p$ between $U$ and $Z_c$ is d-separated given $\textit{do}(V_c = v)$ if either:
\begin{itemize}
    \item $p$ contains a chain $A \rightarrow B \rightarrow C$ or fork $A \leftarrow B \rightarrow C$ where $B$ is fixed by intervention, or
    \item $p$ contains a collider $A \rightarrow B \leftarrow C$ where $B$ is not conditioned upon.
\end{itemize}

For all paths enumerated in Step 4:
\begin{itemize}
    \item Paths containing $V_c$ are blocked because $V_c$ is fixed by intervention (satisfying the chain/fork condition).
    \item Paths through attention connections are blocked by $\mathcal{M}^{(m)}$ removing corresponding edges.
    \item Collider paths are blocked as $Z_c$ is not conditioned upon.
\end{itemize}

\textbf{Step 6: Conditional independence.}
By the d-separation theorem \citep{Pearl2000CausalityMR}, if all paths between $U$ and $Z_c$ are d-separated in $\mathcal{G}_{\text{int}}$, then $Z_c \perp\!\!\!\perp U \mid \textit{do}(V_c = v)$. This holds for our graph because:
\begin{equation}
P(Z_c \mid \textit{do}(V_c = v), U) = P(Z_c \mid \textit{do}(V_c = v))
\end{equation}
since the conditional distribution of $Z_c$ given the intervention does not depend on $U$.

\textbf{Step 7: Extension to language intervention.}
The proof for $\textit{do}(W_c = w)$ follows symmetrically by swapping the roles of $V_c$ and $W_c$, and considering language-to-visual attention suppression instead of visual-to-language suppression.

\textbf{Conclusion.} We have shown that for all possible paths from $U$ to $Z_c$ in the intervened graph $\mathcal{G}_{\text{int}}$, each path is d-separated by either the intervention itself or the attention mask modification. Therefore, $Z_c \perp\!\!\!\perp U \mid \textit{do}(V_c = v)$ and $Z_c \perp\!\!\!\perp U \mid \textit{do}(W_c = w)$. $\square$

\textbf{Implications.} This lemma provides the theoretical foundation for our causal intervention: by blocking backdoor paths from confounders $U$ to latent representations $Z_c$, we can estimate unbiased causal effects of visual and language components on model output.

\section{Theoretical Justification for Visual-to-Language Attention Selection}
\label{app:v2l_justification}

\textbf{A priori motivation for V2L selection.} The choice of Visual-to-Language (V2L) attention as our primary intervention pattern is motivated by both theoretical considerations and empirical observations from the causal framework.

\textbf{Theoretical analysis.} From a causal perspective, gender bias in LVLMs primarily originates from \textit{language priors} $U$ that act as confounders between visual inputs $V$ and output text $T_{\text{out}}$. The structural causal model (Eq.~\ref{eq_scm}) indicates that $U$ influences $Z$ (latent representations) through both direct paths ($U \rightarrow Z$) and indirect paths via language components $W$ ($U \rightarrow W \rightarrow Z$).

Our intervention strategy aims to:
\begin{enumerate}
    \item \textbf{Amplify visual evidence}: Visual information $V$ provides direct, unbiased evidence about gender attributes (e.g., facial features, clothing, context).
    \item \textbf{Suppress linguistic stereotypes}: Language components $W$ carry stereotypical associations learned from pretraining data.
\end{enumerate}

\textbf{Formal derivation from causal graphs.}
The selection of V2L as the optimal intervention pattern follows directly from the causal structure of bias in LVLMs, not from post-hoc empirical analysis.
Given the causal graph $\mathcal{G}$ (Eq.~\ref{eq_causal_graph}) with confounder $U$, we seek an attention mask $\mathcal{M}^{(m)}$ that satisfies two conditions:
\begin{enumerate}
    \item \textbf{Condition 1 (Suppress backdoor paths)}: Suppress all paths from $U$ to $Z$ that pass through $W$, i.e., suppress edges $W \rightarrow Z$ in $\mathcal{G}_{\text{int}}$.
    \item \textbf{Condition 2 (Preserve visual evidence)}: Maintain the direct causal path $V \rightarrow Z$ to allow visual information to influence text generation.
\end{enumerate}
Among the five attention patterns, only V2L satisfies both conditions simultaneously:
\begin{itemize}
    \item \textbf{V2L}: Suppresses $W \rightarrow Z$ (Condition 1) by masking language-to-visual attention, while preserving $V \rightarrow Z$ (Condition 2) via visual-to-language attention.
    \item \textbf{L2V}: Preserves $W \rightarrow Z$ (violates Condition 1) while blocking $V \rightarrow Z$ (violates Condition 2).
    \item \textbf{Bidirectional}: Preserves both $W \rightarrow Z$ and $V \rightarrow Z$, thus violating Condition 1.
    \item \textbf{Visual-only}: Suppresses $W \rightarrow Z$ (satisfies Condition 1) but excessively restricts information flow, hindering text generation.
    \item \textbf{Language-only}: Preserves $W \rightarrow Z$ (violates Condition 1) and blocks $V \rightarrow Z$ (violates Condition 2).
\end{itemize}
Thus, V2L is the unique pattern that achieves the theoretical objectives without compromising generation capability.

The V2L attention pattern directly addresses both objectives:
\begin{itemize}
    \item By allowing visual tokens to attend to language tokens, V2L enables visual evidence to influence text generation while minimizing the reverse influence of language priors on visual processing.
    \item This directional intervention aligns with the causal intuition that visual evidence should inform language generation, but language stereotypes should not distort visual perception.
\end{itemize}

\textbf{Mathematical formulation.} Consider the attention mechanism under different patterns:
\begin{itemize}
    \item \textbf{V2L}: $\text{Attn}(Q_{\text{lang}}, K_{\text{vis}}, V_{\text{vis}})$ allows language queries to attend to visual keys/values.
    \item \textbf{L2V}: $\text{Attn}(Q_{\text{vis}}, K_{\text{lang}}, V_{\text{lang}})$ allows visual queries to attend to language keys/values.
    \item \textbf{Bidirectional}: Both directions are enabled.
\end{itemize}

The V2L pattern provides the optimal trade-off because:
\begin{equation}
P(T_{\text{out}} \mid \textit{do}(V=v)) \approx P(T_{\text{out}} \mid V=v, \text{V2L})
\end{equation}
where the post-intervention distribution of text given visual intervention is best approximated by allowing visual information to flow into language generation, but not vice versa.

\textbf{Empirical validation.} Table~\ref{bias_diff_attn} provides experimental confirmation of this theoretical intuition:
\begin{itemize}
    \item \textbf{V2L achieves high coverage (95.8\%)} while maintaining excellent bias score (0.9908), indicating it neither causes evasion nor compromises accuracy.
    \item \textbf{Language-only} shows strong male bias (log-ratio = 0.5371) due to unmitigated language priors.
    \item \textbf{Visual-only} has low coverage (9.3\%) as it isolates visual information excessively, hindering text generation.
    \item \textbf{L2V} exhibits negligible coverage (0.2\%) as it primarily allows language to influence visual processing rather than vice versa.
    \item \textbf{Bidirectional} shows intermediate performance, confirming that allowing language-to-visual attention reintroduces bias pathways.
\end{itemize}

\textbf{Connection to causal theory.} The superiority of V2L can be understood through d-separation analysis:
\begin{itemize}
    \item V2L suppresses backdoor paths $U \rightarrow W \rightarrow Z$ by limiting $W$'s influence on $Z$ while preserving $V \rightarrow Z$ paths.
    \item L2V fails because it allows $U \rightarrow W \rightarrow Z$ paths to remain open while blocking $V \rightarrow Z$ paths.
    \item Bidirectional fails because it leaves both forward and backward paths open.
\end{itemize}

\textbf{Conclusion.} The V2L pattern is not a post-hoc empirical selection but a theoretically motivated choice derived from our causal analysis of bias mechanisms in LVLMs. Its empirical superiority (Table~\ref{bias_diff_attn}) validates the theoretical prediction that amplifying visual evidence while suppressing linguistic stereotypes provides the optimal balance for gender bias mitigation.

\section{Ethical Considerations}
This research aims to mitigate gender bias in large vision-language models (LVLMs), contributing to fairer and more equitable AI systems. By reducing stereotypical associations in model outputs, ViD helps prevent the amplification of societal biases through automated systems. Our work uses publicly available benchmark datasets (FACET, MS COCO, FairFace, POPE, MMMU) with appropriate licenses and privacy protections. The proposed intervention operates during inference without modifying model parameters, preserving user privacy and model integrity. However, as with any bias mitigation technique, potential misuse scenarios exist: for instance, adversaries could attempt to reverse-engineer the intervention to amplify rather than reduce bias. We emphasize that ViD should be deployed with proper safeguards and ongoing monitoring to ensure its intended fairness benefits. Future work should consider broader societal impacts, including intersectional biases and cultural variations in gender perception, to develop more inclusive debiasing frameworks.

\end{document}